%% file: long_paper_acl_storytelling.tex
\documentclass[11pt,a4paper]{article}
\usepackage[hyperref]{acl2019}
\usepackage{times}
\usepackage{latexsym}
\usepackage{url}

\usepackage{graphicx}
\usepackage{subfig}
\usepackage{wrapfig}
\usepackage{booktabs}
\usepackage{microtype}
\usepackage{amsmath}
\usepackage{multicol}
\usepackage{verbatim}
\usepackage{pifont}
\usepackage{longtable}
\usepackage{enumitem}
\usepackage{rotating}
\usepackage{adjustbox}

\usepackage{array}

\newcolumntype{R}[2]{%
	>{\adjustbox{angle=#1,lap=\width-(#2)}\bgroup}%
	l%
	<{\egroup}%
}
\newcommand*\rot{\multicolumn{1}{R{45}{1em}}}% no optional argument here, please!

\newtheorem{exmp}{Example}[section]

\aclfinalcopy % Uncomment this line for the final submission
\def\aclpaperid{***} %  Enter the acl Paper ID here

\title{Detecting Everyday Scenarios in Narrative Texts}

\author{Lilian D. A. Wanzare\textsuperscript{1}    \hspace{0.75cm} Michael Roth\textsuperscript{2}  \hspace{0.75cm} Manfred Pinkal\textsuperscript{1} \\
         Universit\"{a}t des Saarlandes\textsuperscript{1}     \hspace{0.75cm} 
         	Universit\"{a}t Stuttgart\textsuperscript{2}\\
         \tt \{wanzare,pinkal\}coli.uni-saarland.de  \hspace{0.75cm}  \tt rothml@ims.uni-stuttgart.de}

\date{}

\begin{document}
\maketitle

\begin{abstract}
Script knowledge consists of detailed information on everyday activities. Such information is often taken for granted in text and needs to be inferred by readers. Therefore, script knowledge is a central component to language comprehension. Previous work on representing scripts 
is mostly based on extensive manual work or limited to scenarios that can be found with sufficient redundancy in large corpora. We introduce the task of \textit{scenario detection}, in which we identify references to scripts. In this task, we address a wide range of different scripts (200 scenarios) and we attempt to identify all references to them in a collection of narrative texts. We present a first benchmark data set and a baseline model that tackles scenario detection using techniques from topic segmentation and text classification.

\end{abstract}

\input{introduction}
\input{related_work}

\input{data_annotation}

\input{model}

\input{evaluation}
\input{conclusion}

\section*{Acknowledgments}
This research was funded by the German Research
Foundation (DFG) as part of SFB 1102 ‘Information Density and Linguistic Encoding’.

\bibliography{bibliography}
\bibliographystyle{acl_natbib}
\nopagebreak[4]

\newpage
%~\\
\newpage

\appendix
\section*{Appendix}
\input{segment_annotation_guidelines}

\onecolumn
\input{scenario_list}

\end{document}

%% file: introduction.tex
\section{Introduction}
\label{sec:intro}

According to Grice's  \citeyearpar{grice1975logic} theory of pragmatics, people  tend to omit basic information when participating in a conversation (or writing a story) under the assumption that left out details are already known or can be inferred from commonsense knowledge by the hearer (or reader). Consider the following text fragment about \texttt{eating in a restaurant} from an online blog post: 
 \begin{exmp}
 \label{exmp:intro} (\ldots) we drove to Sham Shui Po and looked for a place to eat. (\ldots) [O]ne of the restaurants was fully seated [so we] chose another. We had 4 dishes---Cow tripe stir fried with shallots, ginger and chili. 1000-year-old-egg with watercress and omelet. Then another kind of tripe and egg---all crispy on the top and soft on the inside. Finally calamari stir fried with rock salt and chili. Washed down with beers and tea at the end. (\ldots)
 \end{exmp}

The text in Example \ref{exmp:intro} obviously talks about a restaurant visit, but it omits many events that are involved while \texttt{eating in a restaurant}, such as \emph{finding a table}, \emph{sitting down}, \emph{ordering food} etc., as well as participants such as %in \texttt{eating in a restaurant}
\emph{the waiter}, \emph{the menu},\emph{the bill}.  A human reader of the story will naturally assume that all these ingredients have their place in the reported event, based on their common-sense knowledge, although the text leaves them completely implicit. For text understanding machines that lack appropriate common-sense knowledge, the implicitness however poses a non-trivial challenge.

Writing and understanding of narrative texts makes particular use of a specific kind of common-sense knowledge, referred to as  \emph{script knowledge} \cite{schank1977}. Script knowledge is about prototypical everyday activity, called  \emph{scenarios}. Given a specific scenario, the associated script knowledge enables us to infer omitted events that happen before and after an explicitly mentioned event, as well as its associated participants. 
In other words, this knowledge can help us obtain more complete text representations, as required for many language comprehension tasks.

There has been some work on script parsing \cite{simon2017scripts,simon2018entailment}, i.e., associating texts with script structure given a specific scenario. 
Unfortunately, only limited previous work exists on determining which scenarios are referred to in a text or text segment (see Section~\ref{sec:background}). To the best of our knowledge, this is the first dataset of narrative texts which have annotations at sentence level according to the scripts they instantiate.

In this paper, we describe first steps towards the automatic detection and labeling of scenario-specific text segments. Our contributions are as follows:
\setlist{itemsep=0.05em}
\begin{itemize} %[noitemsep]
\item We define the task of scenario detection and introduce a benchmark dataset of annotated narrative texts, with segments labeled according to the scripts they instantiate (Section~\ref{sec:data}). To the best of our knowledge, this is the first dataset of its kind. The   corpus   is   publicly   available   for   scientific   research  purposes  at  this  \url{http://www.sfb1102.uni-saarland.de/?page_id=2582.}
\item As a benchmark model for scenario detection, we present a two-stage model that combines established methods from topic segmentation and text classification (Section~\ref{sec:model}). 
\item Finally, we show that the proposed model achieves promising results but also reveals some of the difficulties underlying the task of scenario detection (Section~\ref{sec:exp}). 
\end{itemize}

%% file: related_work.tex
\section{Motivation and Background}

\label{sec:background}

A major line of research has focused on identifying specific events across documents, for example, as part of the Topic Detection and Tracking (TDT) initiative \cite{allan1998topic,allan2012topic}. The main subject of the TDT intiative are instances of world events such as \textit{Cuban Riots in Panama}. In contrast, everyday scenarios and associated sequences of event types, as dealt with in this paper, have so far only been the subject of individual research efforts focusing either on acquiring script knowledge, constructing story corpora, or script-related downstream tasks. Below we describe significant previous work in these areas in more detail.

\paragraph{Script knowledge.}
Scripts are descriptions of prototypical everyday activities such as \texttt{eating in a restaurant} or \texttt{riding a bus} \cite{schank1977}. Different lines of research attempt to acquire script knowledge. Early researchers attempted to handcraft script knowledge \cite{Mueller1999,gordon2001}. Another line of research focuses on the collection of scenario-specific script knowledge in form of event sequence descriptions (ESDs) via crowdsourcing,
\citep{omics2002,omicsGupta:2004,BoyangLi2012,RaisigWHM09,RegneriKP10,wanzare2016descript}). ESDs are sequences of short sentences, in bullet style, describing how a given scenario is typically realized. The top part of Table~\ref{table:script_corpora} summarizes various script knowledge-bases (ESDs). While datasets like OMICS seem large, they focus only on mundane indoor scenarios (e.g. \texttt{open door, switch off lights}).  
  A third line of research tries to leverage existing large text corpora  to induce script-like knowledge about the topics represented in these corpora. For instance,  \citet{ChambersJ08,chambers2009unsupervised, PichottaM14} leverage newswire texts,  \citet{Manshadi2008,gordon_mining_2010,Rudinger1025,webchild14,tandon2017webchild} leverage web articles while
   \citet{ryu2010automatic,Abend-etal:2015,chu2017wikihow} leverage organized procedural knowledge (e.g. from eHow.com, wikiHow.com).

The top part of Table~\ref{table:script_corpora} summarizes various script knowledge-bases.  Our work lies in between both lines of research and may help to connect them: we take an extended set of specific scenarios as a starting point and attempt to identify instances of those scenarios in a large-scale collection of narrative texts.

\paragraph{Textual resources.} Previous work created script-related resources by crowdsourcing stories that instantiate script knowledge of specific scenarios. For example, \newcite{modi16} and \newcite{simon2018MCScript,Ostermann2019MCScript2} asked crowd-workers to write stories that include mundane aspects of scripts ``as if explaining to a child''. 
The collected datasets, \emph{InScript} and \emph{MCScript}, are useful as training instances of narrative texts that refer to scripts. However, the texts are kind of unnatural and atypical because of their explicitness and the requirement to workers to tell a story that is related to one single scenario only.
\newcite{gordon2009} employed statistical text classification in order to collect narrative texts about personal stories. The \emph{Spinn3r}\footnote{http://www.icwsm.org/data/} dataset \cite{Spinn3r_Dataset_2009} contains about 1.5 Million stories.   \emph{Spinn3r} has been used to extract script information \cite[see below]{rahimtoroghi2016SIGDIAL}. 
In this paper, we use the Spinn3r personal stories corpus as a source for our data collection and annotation. The bottom part of Table~\ref{table:script_corpora} summarizes various script-related resources. The large datasets come with no scenarios labels while the crowdsourced datasets only have scenario labels at story level. Our work provides a more fine grained scenario labeling at sentence level.

\paragraph{Script-related tasks.} Several tasks have been proposed that require or test computational models of script knowledge.  For example, \newcite{Kasch:2010} and \newcite{rahimtoroghi2016SIGDIAL} propose and evaluate a method that automatically creates event schemas, extracted from scenario-specific texts.
\newcite{simon2017scripts} attempt to identify and label mentions of events from specific scenarios in corresponding texts. Finally, \newcite{SemEval2018Task11} present an end-to-end evaluation framework that assesses the performance of machine comprehension models using script knowledge. Scenario detection is a prerequisite for tackling such tasks, because the application of script knowledge requires awareness of the scenario a text segment is about.

%% file: data_annotation.tex
\section{Task and Data}
\label{sec:data}

We define scenario detection as the task of identifying segments of a text that are about a specific scenario and classifying these segments accordingly. For the purpose of this task, we view a segment as a consecutive part of text that consists of one or more sentences. Each segment can be assigned none, one, or multiple labels. 
\begin{table}[]
	\scalebox{0.8}{%
		\begin{tabular}{lrrrr}
			%\toprule
			%\toprule
			\multicolumn{3}{l}{\textbf{Scenario ESD collections}} & \rot{\shortstack[l]{\textbf{Scenarios}}}\  & \rot{\shortstack[l]{\textbf{\# ESDs}} } \\ \midrule
			\multicolumn{3}{l}{SMILE  \citep{RegneriKP10}} & 22 & 386 \\
			\multicolumn{3}{l}{Cooking \citep{RegneriPhd}} & 53 & 2500 \\
			\multicolumn{3}{l}{OMICS \citep{omics2002}}& 175 & 9044 \\
			\multicolumn{3}{l}{\citet{RaisigWHM09}} & 30& 450 \\ 
			\multicolumn{3}{l}{\citet{BoyangLi2012}} & 9 & 500  \\
			\multicolumn{3}{l}{DeScript \citep{wanzare2016descript}}&40&4000\\ \bottomrule
			
			\textbf{Story Corpora} & \rot{\shortstack[l]{\textbf{Scenarios}}}  & \rot{\shortstack[l]{\textbf{\# stories} }}  & \rot{\shortstack[l]{\textbf{Classes}}}  &
			\rot{\shortstack[l]{\textbf{Segs.}}}    \\ \midrule
			%Inscript 
			\citet{modi16} &10 & 1000 &\ding{51} &\ding{55}~~   \\ 
			%MCScript 
			\citet{Ostermann2019MCScript2} &200 & 4000 &  \ding{51}&\ding{55}~~ \\ 
			% \citet{Kasch:2010} &&&\\
			 \citet{rahimtoroghi2016SIGDIAL} %(hand labeled) 
			 &2& 660&\ding{51}&\ding{55}~~ \\
			 %ROC-stories 
			 \citet{mostafazadeh-EtAl:2016}& \ding{55} & \url{\~}50000 & \ding{55}& \ding{55}~~ \\ 
	 		%Spinn3r 
			 \citet{gordon2009} &\ding{55}& \url{\~}1.5M &  \ding{55}&\ding{55}~~  \\ 
			 \textbf{This work}  &200&504& \ding{51}&\ding{51}~~ \\
			\bottomrule
		\end{tabular}}
		\caption {Top part shows scenario collections and number of associated event sequence descriptions (ESDs). Bottom part lists story corpora together with the number of stories and different scenarios covered. The last two columns indicate whether the stories are classified and segmented, respectively.} 
		\label{table:script_corpora}
	\end{table}

\paragraph{Scenario labels. }
As a set of target labels, we collected scenarios from all scenario lists available in the literature (see Table~\ref{table:script_corpora}). During revision, we discarded scenarios that are too vague and general (e.g.~\texttt{childhood}) or atomic (e.g.~\texttt{switch on/off lights}), admitting only reasonably structured activities. Based on a sample annotation of Spinn3r stories, we further added 58 new scenarios, e.g. \texttt{attending a court hearing, going skiing}, to increase coverage.  We deliberately included narrowly related scenarios that stand in the relation of specialisation (e.g. \texttt{going shopping} and \texttt{shopping for clothes}, or in a subscript relation (\texttt{flying in an airplane} and \texttt{checking in at the airport}). These cases are challenging to annotators because segments may refer to different scenarios at the same time. 

Although our scenario list is incomplete, it is representative for the structural problems that can occur during annotation. We have scenarios that have varying degrees of complexity and cover a wide range of everyday activities. 
The complete list of scenarios\footnote{The scenario collection was jointly extended together with the authors of MCScript 	\citep{simon2018MCScript,Ostermann2019MCScript2}. The same set was used in building MCScript 2.0 \citep{Ostermann2019MCScript2}} is provided in Appendix \ref{scenario_list}.

\paragraph{Dataset. }As a benchmark dataset, we annotated 504 texts from the Spinn3r corpus. To make sure that our dataset contains a sufficient number of relevant sentences, i.e., sentences that refer to scenarios from our collection, we selected texts that have a high affinity to at least one of these scenarios. We approximate this affinity using a logistic regression model fitted to texts from MCScript, based on LDA topics \cite{blei2003latent} as features to represent a document. 

\subsection{Annotation}

We follow standard methodology for natural language annotation \cite{orelie2012annotation}.  Each text is independently annotated by two annotators, student assistants, who use an agreed upon set of guidelines that is built iteratively together with the annotators. For each text, the students had to identify segments referring to a scenario from the scenario list, and assign scenario labels.
%\footnote{Our annotators are students in Computational Linguistics.} 
If a segment refers to more than one script, they were allowed to assign multiple labels. 
We worked with a total of four student assistants and used the 
Webanno\footnote{https://webanno.github.io/webanno/} annotation tool \citep{de2016web}. 

The annotators labeled 504 documents, consisting of 10,754 sentences. On average, the annotated documents were 35.74 sentences long.  A scenario label could be either one of our 200 scenarios or \emph{None} to capture sentences that do not refer to any of our scenarios. 

%FOLLOWING: MORE CHALLENGES, IMPORTANT GUIDELINES?

\paragraph{Guidelines. } We developed a set of more detailed guidelines for handling different issues related to the segmentation and classification, which is detailed in Appendix \ref{annotation_guidelines}.
%(section\ref{ssec:gold}). 
A major challenge when annotating segments is deciding when to count a sentence as referring to a particular scenario. For the task addressed here, we consider a segment only if it explicitly realizes aspects of script knowledge that go beyond an evoking expression (i.e., more than one event and participant need to be explicitly realized). Example \ref{exmp:minimal_segment} below shows a text segment with minimal scenario information for \texttt{going grocery shopping} with two events mentioned. In Example \ref{exmp:minimal_segment2}, only the evoking expression is mentioned, hence this example is not annotated.

 \begin{exmp}
 \ding{51}\label{exmp:minimal_segment}\texttt{going grocery shopping}\\ ...We also \textbf{stopped at a small shop} near the hotel to \textbf{get some sandwiches} for dinner...
 \end{exmp}
 \begin{exmp}
 \ding{55}\label{exmp:minimal_segment2}\texttt{paying for gas}\\... A customer was heading for the store to \textbf{pay for gas} or whatever,...
 \end{exmp}

\subsection{Statistics}

\paragraph{Agreement. }
\begin{table}[] 
	\centering
		\resizebox{0.47\textwidth}{!}{%
	\begin{tabular}{@{}c|ccc@{}}
		\toprule
		\textbf{Annotators} & 2                     & 3           & 4           \\ \midrule
		%& kappa(raw)            & kappa(raw)  & kappa(raw)  \\
		1         & 0.57 (\textit{0.65}) & 0.63 (\textit{\textbf{0.72}}) & \textbf{0.64} (\textit{0.70}) \\
		2         &                       & 0.62 (\textit{0.71}) & 0.61 (\textit{0.70}) \\
		3         &                       &             & 0.62 (\textit{0.71}) \\ \bottomrule
	\end{tabular}}
	\caption{Kappa (\textit{and raw}) agreement between pairs of annotators on sentence-level scenario labels}
%	\vspace{-1.2em}
	\label{tab:kappa}
\end{table} To measure agreement, we looked at sentence-wise label assignments for each double-annotated text.  We counted agreement if the same scenario label is assigned to a sentence by both annotators.
%WHAT MEANS SIMILAR?
As an indication of chance-corrected agreement, we computed  Kappa scores \cite{cohen1960coefficient}.   A kappa of 1 means that both annotators provided identical (sets of) scenario labels for each sentence. When calculating raw agreements, we counted agreement if there was at least one same scenario label assigned by both annotators. Table \ref{tab:kappa} shows the Kappa and raw (\textit{in italics}) agreements for each pair of annotators.   On average, the Kappa  score was\emph{ 0.61} ranging from 0.57 to 0.64. The average raw agreement score was \emph{0.70} ranging from 0.65 to 0.72. The Kappa value indicates relatively consistent annotations across annotators even though the task was challenging.
% \usepackage{booktabs}

 We used fuzzy matching to calculate agreement in span between segments that overlap by at least one token.
 	Table~\ref{tab:overlap} shows pairwise \% agreement scores between annotators. On average, the annotators achieve  67\% agreement on segment spans. This shows considerable segment overlap when both annotators agreed that a particular scenario is referenced.
 	
 	% \usepackage{booktabs}

 	\begin{table}[]
 		\centering
 		\begin{tabular}{@{}c|ccc@{}}
 			\toprule
 			\textbf{Annotators}	&	2 & 3            & 4                         \\ \midrule
 			%	&              &              &              \\
 			1 & \textbf{78.8}  & 70.6   & \textit{59.3 } \\
 			2 &              & 66.0 & 64.2  \\
 			3 &              &              & 67.0  \\ 
 			\bottomrule
 		\end{tabular}
 		\caption{Relative agreement on segment spans between annotated segments that overlap by at least one token and are assigned the same scenario label}
% 			\vspace{-1.2em}
 		\label{tab:overlap}
 	\end{table}
 	
\paragraph{Analysis.} 

\begin{figure}[]
  	\centering
  	\includegraphics[trim=1 20 40 40,clip,width=0.49\textwidth]{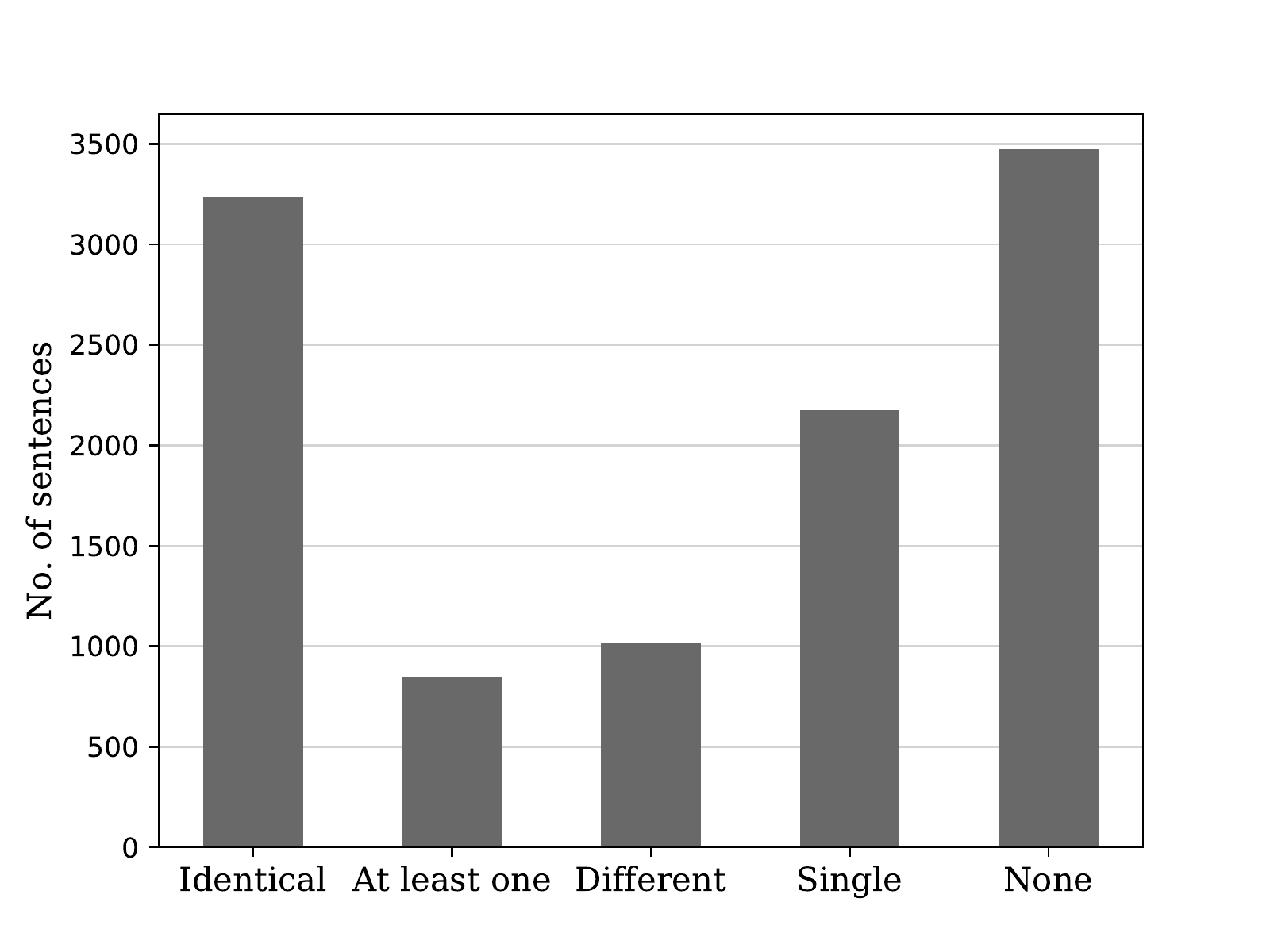} 
  	\caption{Absolute counts on sentence-level annotations that involve the same (\emph{Identical}), overlapping (\emph{At least one}) or disagreed (\emph{Different}) labels; also shown are the number of sentences that received a label by only one annotator (\emph{Single}) or no label at all (\emph{None}).}%
%  		\vspace{-1.2em}
  	\label{fig:segments}%
  \end{figure}

Figure \ref{fig:segments} shows  to what extent the annotators agreed in the scenario labels. The \emph{None} cases accounted for 32\% of the sentences. Our scenario list is by far not complete. Although we selected stories with high affinity to our scenarios, other scenarios (not in our scenario list) may still occur in the stories. 
Sentences referring to other scenarios were annotated as \emph{None} cases. The \emph{None} label was also used to label sentences that described  topics related to but not directly part of the script being referenced. For instance, sentences not part of the narration, but of a different discourse mode (e.g. argumentation, report) or sentences where no specific script events are mentioned\footnote{See examples in Appendix \ref{annotation_guidelines}.}.
About 20\% of the sentences had \emph{Single} annotations where only one annotator indicated that there was a scenario reference. 
47\% of the sentences were assigned some scenario label(s) by both annotators (\emph{Identical, At least one, Different}). Less than 10\% of the sentences had \emph{Different} scenario labels for the case where both annotators assigned scenario labels to a sentence. This occurred frequently with scenarios that are closely related (e.g. \texttt{going to the shopping center, going shopping}) or  scenarios in a sub-scenario relation (e.g. \texttt{flying in a plane, checking in at the airport}) that share script events and participants. In about 7\% of the sentences, both annotators agreed on \emph{At least one} scenario label. The remaining 30\% of the sentences were assigned \emph{Identical} (sets of) scenario labels by both annotators.

\subsection{Adjudication and Gold Standard}

The annotation task is challenging, and so are gold standard creation and adjudication. We combined \emph{ automatic merging} and \emph{manual adjudication}  (by the main author of the paper) as two steps of gold-standard creation, to minimize manual post-processing of the dataset. 

We automatically merged annotated segments that are identical or overlapping and have the same scenario label, thus maximizing segment length. Consider the two annotations shown in Example~\ref{sep2}. One annotator labeled the whole text as \texttt{growing vegetables}, the other one identified the two bold-face sequences as   \texttt{growing vegetables} instances, and left the middle part out. The result of the merger is the maximal \texttt{growing vegetables} chain, i.e., the full text. 
Taking the maximal chain  ensures that all relevant information is included, although the annotators may not have agreed on what is script-relevant. 
%WHAT ISTHE RATIO BEHIND MAXIMATION?
   \begin{exmp} \label{sep2}\texttt{growing vegetables} \\
   	\textbf{The tomato seedlings Mitch planted in the compost box have done really well and we noticed flowers on them today . Hopefully we will get a good }
 It has rained and rained here for the past month so that is doing the garden heaps of good . We bought some organic herbs seedlings recently and now have some thyme , parsley , oregano and mint growing in the garden .\textbf{We also planted some lettuce and a grape vine . We harvested our first crop of sweet potatoes a week or so ago (\ldots)}
  \end{exmp}

The adjudication guidelines were deliberately designed in a way that the adjudicator could not easily overrule the double-annotations. The segmentation could not be changed, and only the labels provided by the annotators were available for labeling. 
 Since segment overlap is handled automatically, manual adjudication must only care about label disagreement: the two main cases are (1) a segment has been labeled by only one annotator and (2)  a segment has been assigned different labels by its two annotators. 
In case (1), the adjudicator had to take a binary decision to accept the labeled segment, or to discard it. In case (2), the adjudicator had three options: to decide for one of the labels or to accept both of them.

%\begin{comment}

\begin{table}[t]
	\centering
		\scalebox{0.95}{%
		\begin{tabular}{@{}lrrr@{}}
			\toprule
			\textbf{Scenario} & \rot{\shortstack[l]{\textbf{\# docs}}} & \rot{\shortstack[l]{\textbf{\# sents.}}} & \rot{\shortstack[l]{\textbf{\# segs}.}} 
			%&\rot{\shortstack[l]{ Avg. seg.\\ length}} 
			\\ \midrule
			eat in a restaurant & ~~~~21 & ~~~387 & ~~~~22 %& 18
			 \\
			go on vacation & 16 & 325 & 17 %& 19 
			\\
			go shopping & 34 & 276 & 35 %& 8
			 \\
			take care of children & 15 & 190 & 19 %& 10
			 \\
			review movies & 8 & 184 & 8 %& 23 
			\\
			%shop for clothes & 11 & 182 & 12 %& 15 \\
%			work in the garden & 13 & 179 & 17 %& 11		\\
			%prepare dinner & 14 & 155 & 17 %& 9 \\
			%play a board game & 8 & 129 & 12 %& 11 \\
			%attend a wedding  & 9 & 125 & 9 %& 14 \\ 
			\multicolumn{4}{c}{\ldots} \\
%			\bottomrule		
%			&  &  &  &  \\
%			\textbf{Avg. frequent scenarios} &  &  &  &  \\ 	\toprule
			taking a bath & 3 & 34 & 6 %& 6 
			\\
			borrow book from library & 3 & 33 & 3 %& 11
			 \\
			mow the lawn & 3 & 33 & 3 %& 11
			 \\
			drive a car & 9 & 32 & 11 %& 3 
			\\
			change a baby diaper & 3 & 32 & 3 %& 11
			 \\
			%make omelette & 3 & 32 & 3 %& 11\\
			%play music in church & 2 & 32 & 3 %& 11 			\\
			%take medicine & 5 & 31 & 6 %& 5 \\
			%get a haircut & 3 & 30 & 3 %& 10 			\\
			%heat food in a microwave & 3 & 30 & 4 %& 8\\
			\multicolumn{4}{c}{\ldots} \\			
%			 \bottomrule
%			&  &  &  &  \\
%			\textbf{Least frequent scenarios} &  &  &  &  \\ 	\toprule
			replace a garbage bag & 1 & 3 & 2 %& 2 
			\\
			unclog the toilet & 1 & 3 & 1 %& 3 
			\\
			wash a cut & 1 & 3 & 1 %& 3
			 \\
			apply band aid & 2 & 2 & 2 %& 1 
			\\
			change batteries in alarm & 1 & 2 & 1 %& 2
			 \\
			%clean a kitchen & 1 & 2 & 1 %& 2\\
			%feed the fish & 1 & 2 & 1 %& 2 			\\
			%set an alarm & 1 & 2 & 1 %& 2 			\\
			%get ready for bed & 1 & 1 & 1 %& 1 			\\
			%set the dining table & 1 & 1 & 1 %& 1 			\\
			 \bottomrule
		\end{tabular}%
	}
	\caption{Distribution of scenario labels over documents (docs), sentences (sents) and segments (segs); the top and bottom parts show the ten most and least frequent labels, respectively. The middle part shows scenario labels that appear at an average frequency. }
	%\vspace{-1.2em}
	\label{table:scenarios_found}
\end{table}
%\end{comment}

\begin{figure}%
	\centering
	%trim={<left> <lower> <right> <upper>}
		\includegraphics[trim=1 12 10 10,clip,width=0.48\textwidth]{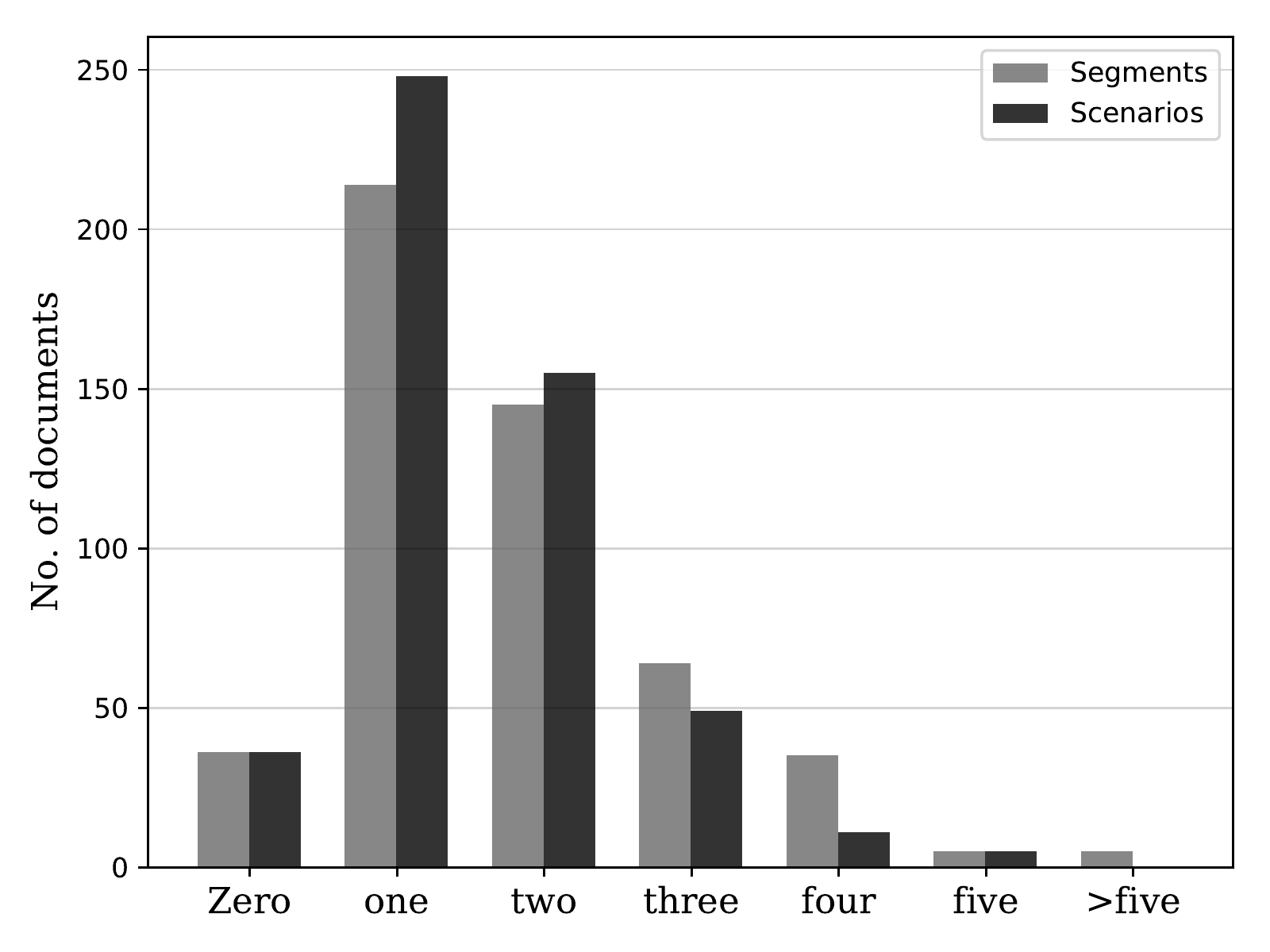} 
	\caption{Segment and scenario distribution per text }%
%		\vspace{-1.2em}
	\label{fig:count_segments}%
\end{figure}

  \paragraph{Gold standard.}The annotation process resulted in 2070 single segment annotations.
  69\% of the single segment annotations were automatically merged to create gold segments. The remaining segments were adjudicated, and relevant segments were added to the gold standard. 
 Our final dataset consists of 7152 sentences (contained in 895 segments) with gold scenario labels. From the 7152 gold sentences, 
 1038 (15\%)
  sentences have more than one scenario label.
181 scenarios (out of 200) occur as gold labels in our dataset, 179 of which are referred to in at least 2 sentences.
Table \ref{table:scenarios_found} shows example  scenarios\footnote{The rest of the scenarios are listed in Appendix \ref{scenario_list}} and the distribution of scenario labels: the number of documents that refer to the given scenario, the number of gold sentences and segments referring to the given scenario, and the average segment length (in sentences) per scenario. 16 scenarios are referred to in more than 100 gold sentences, 105 scenarios in at least 20 gold sentences, 60 scenarios in less than 20 gold sentences. 
Figure \ref{fig:count_segments} shows the distribution of segments and scenario references per text in the gold standard.  	
 On average, there are 1.8 segments per text and 44\% of the texts refer to at least two scenarios.

%% file: model.tex
\section{Benchmark model}
\label{sec:model}
Texts typically consist of different passages that refer to different scenarios. When human hearers or readers come across an expression that evokes a particular script, they try to map verbs or clauses in the subsequent text  to script events, until they face lexical material that is clearly unrelated to the script and may evoke a different scenario. Scenario identification, scenario segmentation, and script parsing are subtasks of story comprehension, which ideally work in close mutual interaction. In this section, we present a model for scenario identification, which is much simpler in several respects: we propose a two-step model consisting of a segmentation and a classification component. For segmentation, we assume that a change in scenario focus can be modeled by a shift in lexical cohesion. We identify segments that might be related to specific scripts or scenarios via topic segmentation, assuming that scenarios can be approximated as distributions over topics. After segmentation, a supervised classifier component is used to predict the scenario label(s) for each of the found segment. Our results show that the script segmentation problem can be solved in principle, and we propose our model as a benchmark model for future work.

\paragraph{Segmentation.}  

The first component of our benchmark model reimplements a state-of-art unsupervised method for topic segmentation, called TopicTiling \citep{riedl2012topictiling}. TopicTiling (TT) 
uses latent topics inferred by a Latent Derichlet Allocation (LDA, \citet{blei2003latent}) model to identify 
segments (i.e., sets of consecutive sentences) referring to similar topics.\footnote{We used the Gensim \cite{rehurek2010software} implementation of LDA.}
The TT segmenter outputs topic boundaries between sentences where there are topic shifts. Boundaries are computed based on coherence scores. Coherence  scores close to 1 indicate significant topic similarity while values close to 0 indicate minimal topic similarity.
A window parameter is used to determine the block size i.e. the number of sentences to the left and right that should be considered when calculating  coherence scores. To discover  segment boundaries, all local minima in the coherence scores are  identified using a depth score \citep{hearst1994multi}.
 A threshold $\mu-\sigma/x$  is used to estimate the number of segments,  where $\mu$ is the mean and $\sigma$ is the standard deviation of the depth scores, and $x$ is a weight parameter for setting the threshold.\footnote{We experimentally set $x$ to 0.1 using our held out development set.} Segment boundaries are placed at positions greater than the threshold.

\paragraph{Classification.} 

We view the scenario classification subtask as a supervised multi-label classification problem. Specifically, we implement a multilayer perceptron classifier in Keras \cite{chollet2015keras} with multiple layers: an input layer with 100 neurons and ReLU activation, followed by an intermediate layer with dropout (0.2), and finally an output layer with sigmoid activations. We optimize a cross-entropy loss using adam. Because multiple labels can be assigned to one segment, we train several one-vs-all classifiers, resulting in one classifier per scenario. 

We also experimented with different features and feature combinations 
to represent text segments: term frequencies weighted by inverted document frequency (\emph{tf.idf}, \citet{salton1986introduction})\footnote{We use SciKit learn \cite{pedregosa2011scikit} to build \emph{tf.idf} representations} 
and  topic features derived from LDA (see above), and we tried to work with word embeddings. We found the performance with \emph{tf.idf} features to be the best. 
%and word embeddings trained with fasttext \cite{joulin2016bag}.

%% file: evaluation.tex
\section{Experiments}\label{sec:exp}

The experiments and results presented in this section are based on  our annotated dataset for scenario detection described in section~\ref{sec:data}. 

\subsection{Experimental setting}
\paragraph{Preprocessing and model details.}
We represent each input to our model as a sequence of lemmatized content words, in particular nouns and verbs (including verb particles). This is achieved by preprocessing each text using Stanford CoreNLP \cite{chen2014fast}.

\paragraph{Segmentation.}Since the segmentation model is unsupervised, we can use all data from both  MCScript and the Spinn3r personal stories corpora to build the LDA model.  
As input to the TopicTiling segmentor, each sentence is represented by a vector in which each component represents the (weight of a) topic from the LDA model 
(i.e. the value of the $i^{th}$ component is the normalized weight of the words in the sentence whose most relevant topic is the $i^{th}$ topic). For the segmentation model, we tune the number of topics (200) and the window size (2) based on an artificial development dataset, created by merging segments from multiple documents from MCScript.

\paragraph{Classification.}We train the scenario classification model on the scenario labels provided in MCScript (one per text).
For training and hyperparameter selection, we split MCScript dataset (see Section~\ref{sec:background}) into a training and development set, as indicated in Table~\ref{tab:exp-data}.  We additionally make use of 18 documents from our scenario detection data (Section~\ref{sec:data}) to tune a classification threshold. The remaining 486 documents are held out exclusively for testing (see Table~\ref{tab:exp-data}). 
Since we train separate classifiers for each scenario (one-vs-all classifiers), we get a probability distribution of how likely a sentence refers to a scenario. We  use entropy to measure the degree of scenario content in the sentences.  Sentences with  entropy values higher than the threshold are considered as not referencing any scenario (\emph{None} cases), while sentences with  lower entropy values reference some scenario. 

\begin{table}[t]
	\centering
	\begin{tabular}{@{}llll@{}}
		\toprule
		\textbf{Dataset} & \textbf{\# train} & \textbf{\# dev} & \textbf{\# test} \\ \midrule
		MCScript & 3492 & 408 & - \\
		Spinn3r (gold) & - & 18 & 486 \\ \bottomrule
	\end{tabular}
	\caption{Datasets (number of documents) used in the experiments}
%	\vspace{-1.2em}
	\label{tab:exp-data}
\end{table}

\paragraph{Baselines.} We experiment with three informed baselines: As a lower bound for the classification task, we compare our model against the baseline \emph{sent\_maj}, which assigns the majority label to all sentences. To assess the utility of segmentation, we compare against two baselines that use our proposed classifier but not the segmentation component: the baseline \emph{sent\_tf.idf} treats each sentence as a separate segment and \emph{random\_tf.idf} splits each document into random segments.

\paragraph{Evaluation.} We evaluate scenario detection performance at the sentence level using micro-average precision, recall and F$_1$-score.  We  consider the top 1 predicted scenario for sentences with only one gold label (including the \emph{None} label), and top \emph{n} scenarios for sentences with \emph{n} gold labels.
For sentences with multiple scenario labels, we take into account partial matches and count each label proportionally. 
Assuming the gold labels are \texttt{washing ones hair} and \texttt{taking a bath}, and the classifier predicts \texttt{taking a bath} and  \texttt{getting ready for bed}.  
\texttt{Taking a bath} is correctly predicted and accounts for $0.5$ true positive (TP) while \texttt{washing ones hair} is incorrectly missed, thus accounts for $0.5$ false negative (FN). \texttt{Getting ready for bed} is incorrectly predicted and accounts for $1$ false positive (FP).  
\begin{comment}
	\begin{exmp}\label{exmp:dis_count}\emph{Counting partial matches}\\ \\
	\textbf{Gold labels}:\emph{[\texttt{washing ones hair}}, \emph{\texttt{taking a bath}]}\\
	\textbf{Predicted labels}:\emph{[\texttt{getting ready for bed}}, \emph{\texttt{taking a bath}]} \\
	
	%{ 'taking a bath': [TP: 0.5], 'washing ones hair': [FN: 0.5], 'getting ready for bed': [FP: 1]}
	\begin{tabular}{@{}l|ccc@{}}
	\textbf{Scenario counts}             & \textbf{TP}   & \textbf{FN}   & \textbf{FP} \\ \midrule
	\emph{\texttt{taking a bath}} &0.5&-& -\\
	\emph{\texttt{washing ones hair}}&-&0.5&-\\
	\emph{\texttt{getting ready for bed}}&-&-&1 \\\bottomrule
	\end{tabular}
	\end{exmp}content...
\end{comment}

We additionally provide separate results of the segmentation component based on standard segmentation evaluation metrics.

\subsection{Results}

\begin{table}[t]
	\centering
	\resizebox{0.4\textwidth}{!}{%
		\begin{tabular}{@{}lccc@{}}
			\toprule
			\textbf{Model}             & \textbf{Precision}   & \textbf{Recall}   & \textbf{F$_1$-score} \\ \midrule

			%sent\_random &  & & 0  \\
			sent\_maj       & 0.08 & 0.05 & 0.06\\ 
			sent\_\emph{tf.idf} &0.24  &0.28 &0.26  \\
			random\_\emph{tf.idf} & 0.32 & 0.45  & 0.37  \\
			
			TT\_\emph{tf.idf} (F\_1)    & 0.36 & \textbf{0.54} & \textbf{0.43} \\
			%TT\_\emph{tf.idf} (Precision) & \textbf{0.60} & 0.42 & \textbf{0.50} \\
			\midrule
			TT\_\emph{tf.idf} (Gold)   & 0.54&0.54 &0.54\\
			\bottomrule
		\end{tabular}%
	}
	\caption{Results for the scenario detection task}
%	\vspace{-1.2em}
	\label{tab:results}
\end{table}

We present the micro-averaged  results for scenario detection in Table~\ref{tab:results}. The \emph{sent\_maj } baseline achieves a F$_1$-score of only 6\%, as the majority class forms only a small part of the dataset (4.7\%). 
Our TT model with \emph{tf.idf} features surpasses both baselines that perform segmentation only naively (26\% F$_1$) or randomly (37\% F$_1$). This result shows that scenario detection works best when using predicted segments that are informative and topically consistent.

We estimated an upper bound for the classifier by taking into account  the predicted segments from the segmentation step, but during evaluation, only considered those sentences with  gold scenario labels  (TT\_\emph{tf.idf} (Gold)), while ignoring the sentences with \emph{None} label.  
We see an improvement in precision (54\%), showing that the classifier  correctly predicts the right scenario label for sentences with gold labels while also including other sentences that may be in topic but not directly referencing a given scenario.

To estimate the performance of the TT segmentor individually, we run TT on an artificial development set, created by merging segments from different scenarios from MCScript. We evaluate the performance of TT by using two standard topic segmentation evaluation metrics, $P_k$ \cite{beeferman1999statistical} and WindowDiff ($WD$, \citet{pevzner2002critique}). Both metrics express the probability of segmentation error, thus lower values indicate better performance. We compute the average performance over several runs. TT attains $P_k$ of 0.28 and $WD$ of 0.28. The low segmentation errors suggest that TT segmentor does a good job in predicting the scenario boundaries.

 \begin{table}[t]
 	\centering
 	\resizebox{0.48\textwidth}{!}{%
 		\begin{tabular}{@{}llcc@{}}
 			\toprule
 			\textbf{True label}       & \textbf{Predicted label}        & \textbf{\# sents.} &\textbf{PMI}\\ \midrule
 			
 			go\_vacation & visit\_sights & 92 & 3.96 \\
 			eat\_restaurant & food\_back & 67 & 4.26 \\
 			work\_garden & grow\_vegetables & 57 & 4.45 \\
 			attend\_wedding & prepare\_wedding & 48 & 4.12 \\
 			eat\_restaurant & dinner\_reservation & 39 & 4.26 \\
 			throw\_party & go\_party & 36 & 4.09 \\
 			shop\_online & order\_ on\_phone & 35 & 3.73 \\
 			%fly & check\_in\_airport & 35 & 5.73 \\
 			%fuel\_car & pay\_gas & 35 & 6.49 \\
 			work\_garden&	planting a tree	&	33&	4.81\\
 			shop\_clothes&	check\_store\_open	&33&	0.00\\
 			play\_video\_games	&learn\_board\_game&	32&	0.00\\
 			
 			%go\_jogging & go\_gym & 32 & 0.00         \\ 
 			\bottomrule
 		\end{tabular}%
 	}
 	\caption{Top 10 misclassified scenario pairs (number of misclassified sentences (\# sents.)) by our approach TT\_\emph{tf.idf} in relation to the PMI  scores for each pair.}
 	\label{tab:confusion}
 	%		\vspace{-1.2em}
 \end{table}
 
 \begin{table}[t]
 	\centering
 	\resizebox{0.46\textwidth}{!}{%
 		\begin{tabular}{@{}lcccc@{}}
 			\toprule
 			\textbf{Scenario} & \textbf{\# sents.}&\textbf{P} & \textbf{R} & \textbf{F$_1$} \\ \midrule
 			go to the dentist & 47&0.90 & 0.96 & 0.93 \\
 			have a barbecue & 43&0.92 & 0.88 & 0.90 \\
 			go to the sauna & 28&0.80 & 0.89 & 0.84 \\
 			make soup & 60&0.81 & 0.87 & 0.84 \\
 			bake a cake &69& 0.71 & 0.97 & 0.82 \\
 			go skiing & 42&0.78 & 0.83 & 0.80 \\
 			attend a court hearing&66 & 0.71 & 0.92 & 0.80 \\
 			clean the floor &6 &1.00 & 0.67 & 0.80 \\
 			take a taxi & 27&0.74 & 0.85 & 0.79 \\
 			attend a church service &60& 0.70 & 0.92 & 0.79 \\ \bottomrule
 		\end{tabular}%
 	}
 	\caption{Top 10 scenario-wise Precision (P), Recall (R) and F$_1$-score (F$_1$) results  using our approach TT\_\emph{tf.idf} and the number of gold sentences (\# sents.) for each scenario. } 
 	\label{tab:scenario-results}
 	%\vspace{-1.2em}
 \end{table}
\subsection{Discussion}

Even for a purpose-built model, scenario detection is a difficult task. This is partly to be expected as the task requires the assignment of one (or more) of 200 possible scenario labels, some of which are hard to distinguish. Many errors are due to misclassifications between scenarios that share script events as well as participants and that are usually mentioned in the same text: for example, \texttt{sending food back in a restaurant} requires and involves participants from \texttt{eating in a restaurant}. 
Table~\ref{tab:confusion} shows the 10 most frequent misclassifications by our best model \emph{TT}\_\emph{tf.idf} (F\_1). These errors account for 16\% of all incorrect label assignments (200 $by$ 200 matrix). The 100 most frequent misclassifications account for 63\% of all incorrect label assignments. 
In a quantitative analysis, we calculated the commonalities between scenarios in terms of the pointwise mutual information (PMI) between scenario labels in the associated stories. We calculated PMI using Equation (\ref{PMI}). The probability of a scenario is given by the document frequency of the scenario divided by the number of documents.  
\begin{equation}\label{PMI}
PMI (S_1,S_2) = log\left(\frac{P(S_1 \wedge S_2 )}{P(S_1)\cdot P(S_2)}\right)
\end{equation} 
Scenarios that tend to co-occur in texts have higher PMI scored.
We observe that the scenario-wise recall and F$_1$-scores of our classifier are negatively correlated with PMI scores (Pearson correlation of $-$0.33 and $-$0.17, respectively). These correlations confirm a greater difficulty in distinguishing between scenarios that are highly related to other scenarios. 

On the positive side, we observe that scenario-wise precision and F$_1$-score are positively correlated with the number of gold sentences annotated with the respective scenario label (Pearson correlation of 0.50 and 0.20, respectively). As one would expect, our approach seems to perform better on scenarios that appear at higher frequency. Table~\ref{tab:scenario-results} shows the 10 scenarios for which our approach achieves the best results.

\paragraph{Scenario approximation using topics. } 
We performed an analyses to qualitatively examine in how far topic distributions, as used in our segmentation model, actually approximate scenarios.
For this analysis, we computed a LDA topic model using only the MCScript dataset. We created \emph{scenario-topics} by looking at all the prevalent topics in documents from a given scenario. Table~\ref{tab:sce-words} shows the top 10 words for each scenario extracted from the \emph{scenario-topics}. As can be seen, the topics capture some of the most relevant words for different scenarios. 

\begin{table}[]
	\centering
	\resizebox{0.46\textwidth}{!}{%
		\begin{tabular}{@{}cccc@{}}
			\toprule
			\textbf{order pizza} & \textbf{laundry} & \textbf{gardening} & \textbf{barbecue} %& \textbf{change light bulb} 
			\\ \midrule
			pizza & clothes & tree & invite %& light
			\\
			order & dryer & plant & guest %& bulb 
			\\
			delivery & laundry & hole & grill %& switch
			\\
			decide & washer & water & friend %& shade
			\\
			place & wash & grow & everyone %& lightbulb
			\\
			deliver & dry & garden & beer %& screw
			\\
			tip & white & dig & barbecue %& remove
			\\
			phone & detergent & dirt & food %& turn 
			\\
			number & start & seed & serve %& fixture 
			\\
			minute & washing & soil & season %& socket 
			\\ \bottomrule
		\end{tabular}}
		\caption{Example top 10 scenario-words}
		\label{tab:sce-words}
%		\vspace{-1.2em}
	\end{table}

%% file: conclusion.tex
\section{Summary}
In this paper we introduced the task of scenario detection and curated a benchmark dataset for automatic scenario segmentation and identification. We proposed a benchmark model that automatically segments and identifies text fragments referring to a given scenario. While our model achieves promising first results, it also revealed some of the difficulties in detecting script references. Script detection is an important first step for large-scale data driven script induction for tasks that require the application of script knowledge. We are hopeful that our data and model will form a useful basis for future work.

%% file: segment_annotation_guidelines.tex
\setcounter{exmp}{2}
\section{Annotation guidelines}\label{annotation_guidelines}
You are presented with several stories. Read each story carefully. 
You are required to highlight segments in the text where any of our scenarios is realized. 
\begin{enumerate} %[label=(\Alph*)]
\item A segment can be a clause, a sentence, several sentences or any combination of sentences and clauses. 
\item Usually segments will cover different parts of the text and be labeled with one scenario label each.
\item A text passage is highlighted as realizing a given scenario only if several scenario elements are addressed or referred to in the text, more than just the evoking expression but some more material e.g at least one event and a participant in that scenario is referred to in the text. (see examples (\ref{exmp1} to \ref{exmp5})). 

\item A text passage referring to one scenario does not necessarily need to be contiguous i.e. the scenario could be referred to in different parts of the same text passage, so the scenario label can occur several times in the text. If the text passages are adjacent, mark the whole span as one segment. (see examples (\ref{exmp6} to \ref{exmp10}))
\item One passage of text can be associated with more than one scenario label.
\begin{itemize}
\item A passage of text associated with two or more related scenarios i.e. scenario that often coincide or occur together. (see example \ref{exmp11}). 
%\item Scenarios  can be nested i.e one scenario can occur as part of another more general scenario.
\item A shorter passage of text referring to a given scenario is nested in a longer passage of text referring to a more general scenario. The nested text passage is therefore associated with both the general and specific scenarios. (see example \ref{exmp12}).
\end{itemize}

\item For a given text passage, if you do not find a full match from the scenario list, but a scenario that is related and similar in structure, you may annotate it.  (see example \ref{exmp13}).

\end{enumerate}

\section*{Rules of thumb for annotation}

\begin{enumerate}

\item  Do not annotate if no progress to events is made i.e. the text just mentions the scenario but no clear script events are addressed.
%When something is mentioned as a function and it does not bring the story forward or its very fine grained or too generic or if it is just a general remark, you do not annotate it,or if it is not a narration but a different discourse mode e.g. argumentation, report.  
\begin{exmp}\label{exmp1}\textcolor{black}{ short text with event and participants addressed}\\
\texttt {\ding{51} feeding a child} \\
... Chloe loves to stand around babbling just generally keeping anyone amused as long as you bribe her with a piece of bread or cheese first.\\

\texttt {\ding{51} going grocery shopping} \\
... but first stopped at a local shop to pick up some cheaper beer . We also stopped at a small shop near the hotel to get some sandwiches for dinner .

\end{exmp}
\begin{exmp}\label{exmp2} \textcolor{black}{scenario is just mentioned}\\
\texttt{\ding{55} cooking pasta}
 And a huge thanks to Megan \& Andrew for a fantastic dinner, especially their first ever fresh pasta making effort of salmon filled ravioli - a big winner.
 
%\\\texttt{\ding{55 } paying bills, \ding{55} settling bank transactions } \\
% I do almost all of my banking and   bill paying online. 

\texttt{\ding{55 } riding on a bus, \ding{55 } flying in a plane } \\
and then catch a bus down to Dublin for my 9:30AM flight  the next morning. 

% \\\texttt{\ding{55} eating in a fast food restaurant, \ding{55} feeding an infant, \ding{55} taking care of children} \\
%\emph{Note:   use \texttt{taking care of children} sparingly as it is does not have real events, use it in cases of baby sitting }

 We decide to stop at at Bob Evan's on the way home and feed the children.
 
\end{exmp}

\begin{exmp}\label{exmp3}\textcolor{black}{scenario is implied but no events are addressed}\\
\texttt{\ding{55} answering the phone } \\
one of the citizens nodded and started talking on her cell phone. Several of the others were also on cell phones

\texttt{\ding{55} taking a photograph  } \\
 Here are some before and after shots of Brandon . The first 3 were all taken this past May . I just took this one a few minutes ago.

 \end{exmp}

\begin{exmp}\label{exmp4}\textcolor{black}{different discourse mode that is not narration e.g. information, argumentative, no specific events are mentioned}\\
\texttt{\ding{55} writing a letter} \\
A long time ago, years before the Internet, I used to write to people from other countries. This people I met through a program called Pen Pal. I would send them my mail address, name, languages I could talk and preferences about my pen pals. Then I would receive a list of names and address and I could start sending them letters.  ...
\end{exmp}

\item When a segment refers to more than one scenario, either related scenarios or scenarios where one is more general than the other, if there is only a weak reference to one of the scenarios, then annotate the text with the scenario having a stronger or more plausible reference.

\begin{exmp}\label{exmp5}one scenario is weakly referenced \\
  \texttt{\ding{51} visiting a doctor, \ding{55} taking medicine} \\ \emph{ taking medicine is weakly referenced}\\ 
  Now another week passes and I get a phone call and am told that the tests showed i had strep so i go in the next day and see the doc and he says that i don 't have strep . ugh what the hell . \emph{This time though they actually give me some antibiotic to help with a few different urinary track infections and other things while doing another blood test and urnine test on me .} \\
 \\\texttt{\ding{51} taking a shower, \ding{55} washing ones hair} \\ \emph{washing ones hair is weakly referenced }\\
I stand under the pressure of the shower , the water hitting my back in fierce beats . I stand and dip my hand back , exposing my delicate throat and neck . \emph{My hair gets soaked and detangles in the water as it flows through my hair , every bead of water putting back the moisture which day to day life rids my hair of . I run my hands through my hair shaking out the water as I bring my head back down to look down towards my feet .} The white marble base of the shower shines back at me from below . My feet covered in water , the water working its way up to my ankles but it never gets there . I find the soap and rub my body all over
\end{exmp}

\item   Sometimes there is a piece of  text intervening two instances (or the same instance) of a scenario, that is not directly part of the scenario that is currently being talked about. We call this a separator. Leave out the separator if it is long or talks about something not related to the scenario being addressed. The separator can be included if it is short, argumentative or a comment, or somehow relates to the scenario being addressed.  When there are multiple adjacent instances of a scenario, annotate them as a single unit.

\begin{exmp}\label{exmp6}\textcolor{black}{two mentions of a scenario annotated as one segment} \\
 \texttt{\ding{51} writing a letter} \\
I asked him about a month ago to write a letter of recommendation for me to help me get a library gig. After bugging him on and off for the past month, as mentioned above, he wrote me about a paragraph. I was sort of pissed as it was quite generic and short. 

I asked for advice, put it off myself for a week and finally wrote the letter of recommendation myself. I had both Evan and Adj. take a look at it- and they both liked my version.

\end{exmp}
\begin{exmp}\label{exmp7}\textcolor{black}{a separator referring to topic related to the current scenario is included} \\
\texttt{\ding{51} writing an exam } \\
 The Basic Science Exam (practice board exam) that took place on Friday April 18 was interesting to say the least. We had 4 hours to complete 200 questions, which will be the approximate time frame for the boards as well. I was completing questions at a good pace for the first 1/3 of the exam, slowed during the second 1/3 and had to rush myself during the last 20 or so questions to complete the exam in time. 
 
  \emph{\ding{51} separator: Starting in May, I am going to start timing myself when I do practice questions so I can get use to pacing. There was a lot of information that was familiar to me on the exam (which is definitely a good thing) but it also showed me that I have a LOT of reviewing to do.}
 
  Monday April 21 was the written exam for ECM. This exam was surprisingly challenging. For me, the most difficult part were reading and interpreting the EKGs. I felt like once I looked at them, everything I knew just fell out of my brain. Fortunately, it was a pass/fail exam and I passed. 

\end{exmp}
\begin{exmp} \label{exmp8}a long separator is excluded\\
\texttt{\ding{51} going to the beach} \\
Today , on the very last day of summer vacation , we finally made it to the beach . Oh , it 's not that we hadn 't been to a beach before . We were on a Lake Michigan beach just last weekend . And we 've stuck our toes in the water at AJ 's and my lake a couple of times . But today , we actually planned to go . We wore our bathing suits and everything . We went with AJ 's friend D , his brother and his mom . 

 \emph{\ding{55} separator: 
D and AJ became friends their very first year of preschool when they were two . They live in the next town over and we don 't see them as often as we would like . It 's not so much the distance , which isn 't far at all , but that the school and athletic schedules are constantly conflicting . But for the first time , they are both going back to school on the same day . So we decided to celebrate the end of summer together . }

\texttt{\ding{51} going to the beach} \\
It nearly looked too cold to go this morning ' the temperature didn 't reach 60 until after 9 :00. The lake water was chilly , too cool for me , but the kids didn 't mind . They splashed and shrieked with laughter and dug in the sand and pointed at the boat that looked like a hot dog and climbed onto the raft and jumped off and had races and splashed some more . D 's mom and I sat in the sun and talked about nothing in particular and waved off seagulls .

\end{exmp}
\begin{exmp}\label{exmp9}a short separator is included\\
\ding{51} \texttt{throwing a party} \\
... My wife planned a surprise party for me at my place in the evening - I was told that we 'd go out and that I was supposed to meet her at Dhobi Ghaut exchange at 7 . 

 \emph{\ding{51} separator: 
But I was getting bored in the office around 5 and thought I 'd go home - when I came home , I surprised her ! }

She was busy blowing balloons , decorating , etc with her friend . I guess I ruined it for her . But the fun part started here - She invited my sister and my cousin ...\\

\ding{51} \texttt{visiting sights} \\
Before getting to the museum we swung by Notre Dame which was very beautiful . I tried taking some pictures inside Notre Dame but I dont think they turned out particularly well . After Notre Dame , Paul decided to show us the Crypte Archeologioue . \\
 \emph{\ding{51} separator: This is apparently French for parking garage there are some excellent pictures on Flickr of our trip there . }
 
Also on the way to the museum we swung by Saint Chapelle which is another church . We didnt go inside this one because we hadnt bought a museum pass yet but we plan to return later on in the trip

\end{exmp}
\item Similarly to intervening text (separator), there may be text before or after that is a motivation, pre or post condition for the applications of the script currently being referred to.  Leave out the text if it is long. The text can be included if it is short, or relates to the scenario being addressed.

\begin{exmp}\label{exmp10}\textcolor{black}{the first one or two sentences introduce the topic}\\
\texttt{\ding{51} getting a haircut } \\
\emph{ I AM , however , upset at the woman who cut his hair recently .} He had an appointment with my stylist (the one he normally goes to ) but I FORGOT about it because I kept thinking that it was a different day than it was . When I called to reschedule , she couldn 't get him in until OCTOBER (?!?!?!) ...

 \texttt{\ding{51} baking a cake}\\
\emph{
I tried out this upside down cake from Bill Grangers , Simply Bill . As I have mentioned before , I love plums am always trying out new recipes featuring them when they are in season .} I didnt read the recipe properly so was surprised when I came to make it that it was actually cooked much in the same way as a tarte tartin , ie making a caramel with the fruit in a frying pan first , then pouring over the cake mixture baking in the frypan in the oven before turning out onto a serving plate , the difference being that it was a cake mixture not pastry ....
\end{exmp}
\item If a text passage refers to several related scenarios,  e.g. "renovating a room" and "painting a wall", "laying flooring in a room", "papering a room"; or "working in the garden" and "growing vegetables", annotate all the related scenarios. 

\begin{exmp} \label{exmp11}\textcolor{black}{segment referring to related scenarios}  \\
\texttt{\ding{51} growing vegetables, \ding{51} working in the garden} \\
The tomato seedlings Mitch planted in the compost box have done really well and we noticed flowers on them today. Hopefully we will get a good crop. It has rained and rained here for the past month so that is doing the garden heaps of good.   We bought some organic herbs seedlings recently and now have some thyme, parsley, oregano and mint growing in the garden. We also planted some lettuce and a grape vine.  ...
\end{exmp}
\item If part of a longer text passage refers to a scenario that is more specific than the scenario currently being talked about, annotate the nested text passage with all referred scenarios. 
\begin{exmp}\label{exmp12} \textcolor{black}{nested segment}\\
\texttt{\ding{51} preparing dinner} \\
 I can remember the recipe, it's pretty adaptable and you can add or substitute the vegetables as you see fit!! One Pot Chicken Casserole 750g chicken thigh meat, cut into big cubes olive oil for frying 1
 
  \texttt{\ding{51} preparing dinner, \ding{51} chopping vegetables} \\
  \emph{large onion, chopped 3 potatoes, waxy is best 3 carrots 4 stalks of celery, chopped 2 cups of chicken stock 2 zucchini, sliced large handful of beans 300 ml cream 1 or 2 tablespoons of wholegrain mustard salt and pepper parsley, chopped }
  
  \#\#42 The potatoes and carrots need to be cut into chunks,. I used chat potatoes which are smaller and cut them in half, but I would probably cut a normal potato into quarters. Heat the  oil; in a large pan and then fry the chicken in batches until it is well browned...
  %Remove from the pan and add the onion and the potato, carrot and celery. After 5 minutes, add the chicken and the stock, put a lid on the pan and bring to the boil. Turn down to a simmer and cook for about an hour, stirring and checking it 2 or 3 times. Mix the cream and the mustard and add along with the beans and zucchin. Cook for a further 20-30 minutes uncovered: the sauce should thicken slightly and the vegetables should all be cooked through. Finally season with salt and pepper and mix in some parsley. Serve with rice or pasta. I added some mushrooms and garlic to the casserole. It freezes very well.

\end{exmp}

\item If you do not find a full match for a text segment in the scenario list, but a scenario that is related and similar in its structure, you may annotate it. 
%Annotate a segment if the topic in the segment share the same sense or is of the same spirit as one of our scenarios
\begin{exmp}\label{exmp13} topic similarity
\begin{itemize}
\item Same structure in scenario – e.g. going fishing for leisure or for work, share the same core events in going fishing
\item  Baking something with flour (baking a cake, baking Blondies, )
\end{itemize}

\end{exmp}
\end{enumerate}

%% file: scenario_list.tex
\section{List of Scenarios}
\label{scenario_list}
% \usepackage{booktabs}
% \usepackage{longtable}
% Note: It may be necessary to compile the document several times to get a multi-page table to line up properly
\footnotesize
\begin{longtable}[c]{@{}llllllllll@{}}
\toprule & scenario & \# docs & \# sents. & \# segs. &  & scenario & \# docs & \# sents. & \# segs. \\
 \midrule
%\endfirsthead
\endhead
%\bottomrule
%\endfoot
%\endlastfoot
%

1 & eating in a restaurant & 21 & 387 & 22 & 101 & receiving a letter & 5 & 27 & 5 \\
2 & going on vacation & 16 & 325 & 17 & 102 & taking a shower & 4 & 27 & 4 \\
3 & going shopping & 34 & 276 & 35 & 103 & taking a taxi & 4 & 27 & 6 \\
4 & taking care of children & 15 & 190 & 19 & 104 & going to the playground & 3 & 25 & 4 \\
5 & reviewing movies & 8 & 184 & 8 & 105 & taking a photograph & 5 & 25 & 6 \\
6 & shopping for clothes & 11 & 182 & 12 & 106 & going on a date & 3 & 24 & 3 \\
7 & working in the garden & 13 & 179 & 17 & 107 & making a bonfire & 2 & 23 & 3 \\
8 & preparing dinner & 14 & 155 & 17 & 108 & renting a movie & 3 & 23 & 3 \\
9 & playing a board game & 8 & 129 & 12 & 109 & buying a house & 2 & 22 & 3 \\
10 & attend a wedding ceremony & 9 & 125 & 9 & 110 & designing t-shirts & 2 & 22 & 3 \\
11 & playing video games & 6 & 124 & 7 & 111 & doing online banking & 3 & 22 & 3 \\
12 & throwing a party & 10 & 123 & 12 & 112 & planting flowers & 4 & 22 & 4 \\
13 & eat in a fast food restaurant & 9 & 113 & 12 & 113 & taking out the garbage & 4 & 22 & 4 \\
14 & adopting a pet & 7 & 111 & 7 & 114 & brushing teeth & 3 & 21 & 4 \\
15 & taking a child to bed & 9 & 108 & 11 & 115 & changing bed sheets & 3 & 21 & 4 \\
16 & shopping online & 7 & 102 & 9 & 116 & going bowling & 2 & 21 & 2 \\
17 & going on a bike tour & 6 & 93 & 6 & 117 & going for a walk & 4 & 21 & 4 \\
18 & playing tennis & 5 & 91 & 7 & 118 & making coffee & 2 & 21 & 4 \\
19 & renovating a room & 9 & 87 & 10 & 119 & serving a drink & 5 & 20 & 6 \\
20 & growing vegetables & 7 & 82 & 8 & 120 & taking children to school & 3 & 20 & 3 \\
21 & listening to music & 8 & 81 & 11 & 121 & taking the underground & 2 & 20 & 2 \\
22 & sewing clothes & 6 & 79 & 8 & 122 & feeding a cat & 4 & 19 & 5 \\
23 & training a dog & 3 & 79 & 3 & 123 & going to a party & 5 & 19 & 6 \\
24 & moving into a new flat & 8 & 78 & 9 & 124 & ironing laundry & 2 & 19 & 2 \\
25 & answering the phone & 11 & 75 & 12 & 125 & making tea & 3 & 18 & 3 \\
26 & going to a concert & 5 & 74 & 5 & 126 & sending a fax & 3 & 18 & 3 \\
27 & looking for a job & 5 & 74 & 5 & 127 & sending party invitations & 3 & 18 & 3 \\
28 & visiting relatives & 12 & 73 & 13 & 128 & planting a tree & 3 & 17 & 3 \\
29 & checking in at an airport & 5 & 71 & 5 & 129 & set up presentation equipment & 2 & 17 & 2 \\
30 & making a camping trip & 5 & 71 & 5 & 130 & visiting a museum & 2 & 17 & 2 \\
31 & painting a wall & 8 & 71 & 10 & 131 & calling 911 & 2 & 16 & 2 \\
32 & planning a holiday trip & 12 & 71 & 13 & 132 & changing a light bulb & 3 & 16 & 4 \\
33 & baking a cake & 3 & 69 & 5 & 133 & making toasted bread & 1 & 16 & 1 \\
34 & going to the gym & 6 & 69 & 7 & 134 & playing a song & 2 & 16 & 2 \\
35 & attending a court hearing & 3 & 66 & 4 & 135 & washing clothes & 3 & 16 & 3 \\
36 & going to the theater & 6 & 66 & 6 & 136 & putting up a painting & 2 & 15 & 2 \\
37 & going to a pub & 4 & 65 & 4 & 137 & serving a meal & 5 & 15 & 5 \\
38 & playing football & 3 & 65 & 5 & 138 & washing dishes & 3 & 15 & 3 \\
39 & going to a funeral & 5 & 64 & 5 & 139 & cooking pasta & 2 & 14 & 2 \\
40 & visiting a doctor & 7 & 64 & 11 & 140 & moving furniture & 4 & 14 & 4 \\
41 & paying with a credit card & 6 & 63 & 6 & 141 & putting a poster on the wall & 2 & 13 & 2 \\
42 & settling bank transactions & 5 & 63 & 6 & 142 & cleaning up toys & 1 & 12 & 2 \\
43 & paying bills & 6 & 62 & 6 & 143 & preparing a picnic & 2 & 12 & 2 \\
44 & taking a swimming class & 3 & 62 & 4 & 144 & repairing a bicycle & 2 & 12 & 2 \\
45 & looking for a flat & 6 & 61 & 6 & 145 & cooking meat & 4 & 11 & 5 \\
46 & attend a church service & 3 & 60 & 3 & 146 & drying clothes & 3 & 11 & 3 \\
47 & making soup & 3 & 60 & 4 & 147 & give a medicine to someone & 3 & 11 & 3 \\
48 & flying in a plane & 5 & 57 & 6 & 148 & feeding an infant & 4 & 10 & 4 \\
49 & going grocery shopping & 13 & 57 & 13 & 149 & telling a story & 2 & 10 & 2 \\
50 & walking a dog & 5 & 57 & 6 & 150 & unloading the dishwasher & 1 & 10 & 1 \\
51 & go to the swimming pool & 5 & 56 & 5 & 151 & putting away groceries & 3 & 9 & 3 \\
52 & preparing a wedding & 3 & 56 & 3 & 152 & deciding on a movie & 1 & 7 & 1 \\
53 & writing a letter & 5 & 54 & 7 & 153 & going to a shopping centre & 1 & 7 & 1 \\
54 & buy from a vending machine & 3 & 53 & 3 & 154 & loading the dishwasher & 2 & 7 & 4 \\
55 & attending a job interview & 3 & 52 & 4 & 155 & making a bed & 1 & 7 & 1 \\
56 & visiting sights & 9 & 52 & 13 & 156 & making a dinner reservation & 1 & 7 & 2 \\
57 & attend a football match & 4 & 51 & 6 & 157 & making scrambled eggs & 1 & 7 & 1 \\
58 & cleaning up a flat & 6 & 51 & 6 & 158 & playing piano & 2 & 7 & 2 \\
59 & washing ones hair & 6 & 49 & 6 & 159 & wrapping a gift & 1 & 7 & 1 \\
60 & writing an exam & 5 & 49 & 8 & 160 & chopping vegetables & 3 & 6 & 3 \\
61 & watching a tennis match & 3 & 48 & 3 & 161 & cleaning the floor & 1 & 6 & 2 \\
62 & going to the dentist & 3 & 47 & 5 & 162 & getting the newspaper & 1 & 6 & 2 \\
63 & making a sandwich & 4 & 47 & 5 & 163 & making fresh orange juice & 1 & 6 & 1 \\
64 & playing golf & 3 & 47 & 3 & 164 & checking if a store is open & 2 & 5 & 3 \\
65 & taking a driving lesson & 2 & 44 & 2 & 165 & heating food on kitchen gas & 1 & 4 & 2 \\
66 & going fishing & 4 & 43 & 4 & 166 & locking up the house & 2 & 4 & 2 \\
67 & having a barbecue & 4 & 43 & 5 & 167 & cleaning the bathroom & 2 & 3 & 2 \\
68 & riding on a bus & 6 & 43 & 7 & 168 & mailing a letter & 1 & 3 & 1 \\
69 & going on a train & 4 & 42 & 6 & 169 & making a hot dog & 1 & 3 & 1 \\
70 & going skiing & 2 & 42 & 2 & 170 & playing a movie & 1 & 3 & 1 \\
71 & packing a suitcase & 5 & 42 & 6 & 171 & remove and replace a garbage bag & 1 & 3 & 2 \\
72 & vacuuming the carpet & 3 & 41 & 3 & 172 & taking copies & 2 & 3 & 2 \\
73 & order something on the phone & 6 & 40 & 8 & 173 & unclog the toilet & 1 & 3 & 1 \\
74 & ordering a pizza & 3 & 39 & 4 & 174 & washing a cut & 1 & 3 & 1 \\
75 & going to work & 3 & 38 & 3 & 175 & applying band aid & 2 & 2 & 2 \\
76 & doing laundry & 4 & 37 & 5 & 176 & change batteries in an alarm clock & 1 & 2 & 1 \\
77 & cooking fish & 3 & 36 & 5 & 177 & cleaning a kitchen & 1 & 2 & 1 \\
78 & learning a board game & 1 & 36 & 1 & 178 & feeding the fish & 1 & 2 & 1 \\
79 & fueling a car & 3 & 35 & 3 & 179 & setting an alarm & 1 & 2 & 1 \\
80 & going dancing & 3 & 35 & 4 & 180 & getting ready for bed & 1 & 1 & 1 \\
81 & laying flooring in a room & 4 & 35 & 4 & 181 & setting the dining table & 1 & 1 & 1 \\
82 & making breakfast & 2 & 35 & 2 & 182 & changing batteries in a camera & 0 & 0 & 0 \\
83 & paying for gas & 3 & 34 & 3 & 183 & buying a tree & 0 & 0 & 0 \\
84 & taking a bath & 3 & 34 & 6 & 184 & papering a room & 0 & 0 & 0 \\
85 & visiting the beach & 4 & 34 & 4 & 185 & cutting your own hair & 0 & 0 & 0 \\
86 & borrow book from the library & 3 & 33 & 3 & 186 & water indoor plants & 0 & 0 & 0 \\
87 & mowing the lawn & 3 & 33 & 3 & 187 & organize a board game evening & 0 & 0 & 0 \\
88 & changing a baby diaper & 3 & 32 & 3 & 188 & cleaning the shower & 0 & 0 & 0 \\
89 & driving a car & 9 & 32 & 11 & 189 & canceling a party & 0 & 0 & 0 \\
90 & making omelette & 3 & 32 & 3 & 190 & cooking rice & 0 & 0 & 0 \\
91 & playing music in church & 2 & 32 & 3 & 191 & buying a DVD player & 0 & 0 & 0 \\
92 & taking medicine & 5 & 31 & 6 & 192 & folding clothes & 0 & 0 & 0 \\
93 & getting a haircut & 3 & 30 & 3 & 193 & buying a birthday present & 0 & 0 & 0 \\
94 & heating food in a microwave & 3 & 30 & 4 & 194 & Answering the doorbell & 0 & 0 & 0 \\
95 & making a mixed salad & 3 & 30 & 3 & 195 & cleaning the table & 0 & 0 & 0 \\
96 & going jogging & 2 & 28 & 2 & 196 & boiling milk & 0 & 0 & 0 \\
97 & going to the sauna & 3 & 28 & 3 & 197 & sewing a button & 0 & 0 & 0 \\
98 & paying taxes & 2 & 28 & 2 & 198 & reading a story to a child & 0 & 0 & 0 \\
99 & send food back (in a restaurant) & 2 & 28 & 3 & 199 & making a shopping list & 0 & 0 & 0 \\
100 & making a flight reservation & 2 & 27 & 3 & 200 & emptying the kitchen sink & 0 & 0 & 0 \\* \bottomrule
%\caption{List of scenarios}
\\
\end{longtable}

%% file: long_paper_acl_storytelling.bbl
\begin{thebibliography}{48}
\expandafter\ifx\csname natexlab\endcsname\relax\def\natexlab#1{#1}\fi

\bibitem[{Abend et~al.(2015)Abend, Cohen, and Steedman}]{Abend-etal:2015}
Omri Abend, Shay~B Cohen, and Mark Steedman. 2015.
\newblock Lexical event ordering with an edge-factored model.
\newblock In \emph{Proceedings of the 2015 Conference of the North American
  Chapter of the Association for Computational Linguistics: Human Language
  Technologies}, pages 1161--1171.

\bibitem[{Allan(2012)}]{allan2012topic}
James Allan. 2012.
\newblock \emph{Topic detection and tracking: event-based information
  organization}, volume~12.
\newblock Springer Science \& Business Media.

\bibitem[{Allan et~al.(1998)Allan, Carbonell, Doddington, Yamron, and
  Yang}]{allan1998topic}
James Allan, Jaime~G Carbonell, George Doddington, Jonathan Yamron, and Yiming
  Yang. 1998.
\newblock Topic detection and tracking pilot study final report.

\bibitem[{Beeferman et~al.(1999)Beeferman, Berger, and
  Lafferty}]{beeferman1999statistical}
Doug Beeferman, Adam Berger, and John Lafferty. 1999.
\newblock Statistical models for text segmentation.
\newblock \emph{Machine learning}, 34(1-3):177--210.

\bibitem[{Blei et~al.(2003)Blei, Ng, and Jordan}]{blei2003latent}
David~M. Blei, Andrew~Y. Ng, and Michael~I. Jordan. 2003.
\newblock Latent dirichlet allocation.
\newblock \emph{Journal of machine Learning research}, 3(Jan):993--1022.

\bibitem[{Burton et~al.(2009)Burton, Java, and Soboroff}]{Spinn3r_Dataset_2009}
Kevin Burton, Akshay Java, and Ian Soboroff. 2009.
\newblock {The ICWSM 2009 Spinn3r Dataset}.
\newblock In \emph{Third Annual Conference on Weblogs and Social Media (ICWSM
  2009)}, San Jose, CA. AAAI.

\bibitem[{de~Castilho et~al.(2016)de~Castilho, Mujdricza-Maydt, Yimam,
  Hartmann, Gurevych, Frank, and Biemann}]{de2016web}
Richard~Eckart de~Castilho, Eva Mujdricza-Maydt, Seid~Muhie Yimam, Silvana
  Hartmann, Iryna Gurevych, Anette Frank, and Chris Biemann. 2016.
\newblock A web-based tool for the integrated annotation of semantic and
  syntactic structures.
\newblock In \emph{Proceedings of the Workshop on Language Technology Resources
  and Tools for Digital Humanities (LT4DH)}, pages 76--84.

\bibitem[{Chambers and Jurafsky(2008)}]{ChambersJ08}
Nathanael Chambers and Daniel Jurafsky. 2008.
\newblock Unsupervised learning of narrative event chains.
\newblock In \emph{{ACL} 2008, Proceedings of the 46th Annual Meeting of the
  Association for Computational Linguistics, June 15-20, 2008, Columbus, Ohio,
  {USA}}, pages 789--797.

\bibitem[{Chambers and Jurafsky(2009)}]{chambers2009unsupervised}
Nathanael Chambers and Daniel Jurafsky. 2009.
\newblock Unsupervised learning of narrative schemas and their participants.
\newblock In \emph{Proceedings of 47th Annual Meeting of the ACL and the 4th
  IJCNLP of the AFNLP}, pages 602--610, Suntec, Singapore.

\bibitem[{Chen and Manning(2014)}]{chen2014fast}
Danqi Chen and Christopher Manning. 2014.
\newblock A fast and accurate dependency parser using neural networks.
\newblock In \emph{Proceedings of the 2014 conference on empirical methods in
  natural language processing (EMNLP)}, pages 740--750.

\bibitem[{Chollet et~al.(2015)}]{chollet2015keras}
Fran{\c{c}}ois Chollet et~al. 2015.
\newblock Keras.
\newblock https://github.com/fchollet/keras.

\bibitem[{Chu et~al.(2017)Chu, Tandon, and Weikum}]{chu2017wikihow}
Cuong~Xuan Chu, Niket Tandon, and Gerhard Weikum. 2017.
\newblock Distilling task knowledge from how-to communities.
\newblock In \emph{Proceedings of the 26th International Conference on World
  Wide Web}, WWW '17, pages 805--814, Republic and Canton of Geneva,
  Switzerland. International World Wide Web Conferences Steering Committee.

\bibitem[{Cohen(1960)}]{cohen1960coefficient}
Jacob Cohen. 1960.
\newblock A coefficient of agreement for nominal scales.
\newblock \emph{Educational and psychological measurement}, 20(1):37--46.

\bibitem[{Gordon and Swanson(2009)}]{gordon2009}
Andrew Gordon and Reid Swanson. 2009.
\newblock Identifying personal stories in millions of weblog entries.
\newblock In \emph{International Conference on Weblogs and Social Media, Data
  Challenge Workshop, May 20, San Jose, CA}.

\bibitem[{Gordon(2001)}]{gordon2001}
Andrew~S. Gordon. 2001.
\newblock Browsing image collections with representations of common-sense
  activities.
\newblock \emph{Journal of the American Society for Information Science and
  Technology.}, 52:925.

\bibitem[{Gordon(2010)}]{gordon_mining_2010}
Andrew~S. Gordon. 2010.
\newblock Mining commonsense knowledge from personal stories in internet
  weblogs.
\newblock In \emph{Proceedings of the {First} {Workshop} on {Automated}
  {Knowledge} {Base} {Construction}}, Grenoble, France.

\bibitem[{Grice(1975)}]{grice1975logic}
Herbert~Paul Grice. 1975.
\newblock Logic and conversation.
\newblock In Peter Cole and Jerry~L. Morgan, editors, \emph{Syntax and
  Semantics: Vol. 3: Speech Acts}, pages 41--58. Academic Press, New York.

\bibitem[{Gupta and Kochenderfer(2004)}]{omicsGupta:2004}
Rakesh Gupta and Mykel~J. Kochenderfer. 2004.
\newblock Common sense data acquisition for indoor mobile robots.
\newblock In \emph{Proceedings of the 19th National Conference on Artifical
  Intelligence}, AAAI'04, pages 605--610. AAAI Press.

\bibitem[{Hearst(1994)}]{hearst1994multi}
Marti~A Hearst. 1994.
\newblock Multi-paragraph segmentation of expository text.
\newblock In \emph{Proceedings of the 32nd annual meeting on Association for
  Computational Linguistics}, pages 9--16. Association for Computational
  Linguistics.

\bibitem[{Kasch and Oates(2010)}]{Kasch:2010}
Niels Kasch and Tim Oates. 2010.
\newblock Mining script-like structures from the web.
\newblock In \emph{Proceedings of the NAACL HLT 2010 First International
  Workshop on Formalisms and Methodology for Learning by Reading}, FAM-LbR '10,
  pages 34--42, Stroudsburg, PA, USA. Association for Computational
  Linguistics.

\bibitem[{Li et~al.(2012)Li, Lee-Urban, Appling, and Riedl}]{BoyangLi2012}
Boyang Li, Stephen Lee-Urban, D.~Scott Appling, and Mark~O. Riedl. 2012.
\newblock Crowdsourcing narrative intelligence.
\newblock volume vol. 2. Advances in Cognitive Systems.

\bibitem[{Manshadi et~al.(2008)Manshadi, Swanson, and Gordon}]{Manshadi2008}
M~Manshadi, R~Swanson, and AS~Gordon. 2008.
\newblock Learning a probabilistic model of event sequences from internet
  weblog stories.
\newblock FLAIRS Conference.

\bibitem[{Modi et~al.(2016)Modi, Anikina, Ostermann, and Pinkal}]{modi16}
Ashutosh Modi, Tatjana Anikina, Simon Ostermann, and Manfred Pinkal. 2016.
\newblock Inscript: Narrative texts annotated with script information.
\newblock In \emph{Proceedings of the 10th International Conference on Language
  Resources and Evaluation (LREC 16)}, Portoro\u{z}, Slovenia.

\bibitem[{Mostafazadeh et~al.(2016)Mostafazadeh, Chambers, He, Parikh, Batra,
  Vanderwende, Kohli, and Allen}]{mostafazadeh-EtAl:2016}
Nasrin Mostafazadeh, Nathanael Chambers, Xiaodong He, Devi Parikh, Dhruv Batra,
  Lucy Vanderwende, Pushmeet Kohli, and James Allen. 2016.
\newblock A corpus and cloze evaluation for deeper understanding of commonsense
  stories.
\newblock In \emph{Proceedings of the 2016 Conference of the North American
  Chapter of the Association for Computational Linguistics: Human Language
  Technologies}, pages 839--849, San Diego, California. Association for
  Computational Linguistics.

\bibitem[{Mueller(1999)}]{Mueller1999}
Erik~T. Mueller. 1999.
\newblock A database and lexicon of scripts for thoughttreasure.
\newblock \emph{CoRR}, cs.AI/0003004.

\bibitem[{Ostermann et~al.(2018{\natexlab{a}})Ostermann, Modi, Roth, Thater,
  and Pinkal}]{simon2018MCScript}
Simon Ostermann, Ashutosh Modi, Michael Roth, Stefan Thater, and Manfred
  Pinkal. 2018{\natexlab{a}}.
\newblock {MCScript: A Novel Dataset for Assessing Machine Comprehension Using
  Script Knowledge}.
\newblock In \emph{Proceedings of the 11th International Conference on Language
  Resources and Evaluation (LREC 2018)}, Miyazaki, Japan.

\bibitem[{Ostermann et~al.(2018{\natexlab{b}})Ostermann, Roth, Modi, Thater,
  and Pinkal}]{SemEval2018Task11}
Simon Ostermann, Michael Roth, Ashutosh Modi, Stefan Thater, and Manfred
  Pinkal. 2018{\natexlab{b}}.
\newblock {SemEval-2018 Task 11: Machine Comprehension using Commonsense
  Knowledge}.
\newblock In \emph{Proceedings of International Workshop on Semantic Evaluation
  (SemEval-2018)}, New Orleans, LA, USA.

\bibitem[{Ostermann et~al.(2019)Ostermann, Roth, and
  Pinkal}]{Ostermann2019MCScript2}
Simon Ostermann, Michael Roth, and Manfred Pinkal. 2019.
\newblock {MCScript2.0: A Machine Comprehension Corpus Focused on Script Events
  and Participants}.
\newblock \emph{Proceedings of the 8th Joint Conference on Lexical and
  Computational Semantics (*SEM 2019)}.

\bibitem[{Ostermann et~al.(2017)Ostermann, Roth, Thater, and
  Pinkal}]{simon2017scripts}
Simon Ostermann, Michael Roth, Stefan Thater, and Manfred Pinkal. 2017.
\newblock Aligning script events with narrative texts.
\newblock \emph{Proceedings of *SEM 2017}.

\bibitem[{Ostermann et~al.(2018{\natexlab{c}})Ostermann, Seitz, Thater, and
  Pinkal}]{simon2018entailment}
Simon Ostermann, Hannah Seitz, Stefan Thater, and Manfred Pinkal.
  2018{\natexlab{c}}.
\newblock {Mapping Texts to Scripts: An Entailment Study}.
\newblock In \emph{Proceedings of the Eleventh International Conference on
  Language Resources and Evaluation (LREC 2018)}, Miyazaki, Japan. European
  Language Resources Association (ELRA).

\bibitem[{Pedregosa et~al.(2011)Pedregosa, Varoquaux, Gramfort, Michel,
  Thirion, Grisel, Blondel, Prettenhofer, Weiss, Dubourg
  et~al.}]{pedregosa2011scikit}
Fabian Pedregosa, Ga{\"e}l Varoquaux, Alexandre Gramfort, Vincent Michel,
  Bertrand Thirion, Olivier Grisel, Mathieu Blondel, Peter Prettenhofer, Ron
  Weiss, Vincent Dubourg, et~al. 2011.
\newblock {Scikit-learn: Machine learning in Python}.
\newblock \emph{Journal of machine learning research}, 12(Oct):2825--2830.

\bibitem[{Pevzner and Hearst(2002)}]{pevzner2002critique}
Lev Pevzner and Marti~A Hearst. 2002.
\newblock A critique and improvement of an evaluation metric for text
  segmentation.
\newblock \emph{Computational Linguistics}, 28(1):19--36.

\bibitem[{Pichotta and Mooney(2014)}]{PichottaM14}
Karl Pichotta and Raymond~J. Mooney. 2014.
\newblock {Statistical Script Learning with Multi-Argument Events}.
\newblock In \emph{Proceedings of the 14th Conference of the European Chapter
  of the Association for Computational Linguistics (EACL 2014)}, pages
  220--229, Gothenburg, Sweden.

\bibitem[{Pustejovsky and Stubbs(2012)}]{orelie2012annotation}
James Pustejovsky and Amber Stubbs. 2012.
\newblock \emph{Natural Language Annotation for Machine Learning - a Guide to
  Corpus-Building for Applications}.
\newblock O'Reilly.

\bibitem[{Rahimtoroghi et~al.(2016)Rahimtoroghi, Hernandez, and
  Walker}]{rahimtoroghi2016SIGDIAL}
Elahe Rahimtoroghi, Ernesto Hernandez, and Marilyn Walker. 2016.
\newblock Learning fine-grained knowledge about contingent relations between
  everyday events.
\newblock In \emph{Proceedings of the 17th Annual Meeting of the Special
  Interest Group on Discourse and Dialogue}, pages 350--359.

\bibitem[{Raisig et~al.(2009)Raisig, Welke, Hagendorf, and der
  Meer}]{RaisigWHM09}
Susanne Raisig, Tinka Welke, Herbert Hagendorf, and Elke~Van der Meer. 2009.
\newblock Insights into knowledge representation: The influence of amodal and
  perceptual variables on event knowledge retrieval from memory.
\newblock \emph{Cognitive Science}, 33(7):1252--1266.

\bibitem[{Regneri(2013)}]{RegneriPhd}
Michaela Regneri. 2013.
\newblock \emph{Event Structures in Knowledge, Pictures and Text}.
\newblock Ph.D. thesis, Saarland University.

\bibitem[{Regneri et~al.(2010)Regneri, Koller, and Pinkal}]{RegneriKP10}
Michaela Regneri, Alexander Koller, and Manfred Pinkal. 2010.
\newblock Learning script knowledge with web experiments.
\newblock In \emph{Proceedings of the 48th Annual Meeting of the Association
  for Computational Linguistics}, pages 979--988, Uppsala, Sweden.

\bibitem[{Rehurek and Sojka(2010)}]{rehurek2010software}
Radim Rehurek and Petr Sojka. 2010.
\newblock Software framework for topic modelling with large corpora.
\newblock In \emph{In Proceedings of the LREC 2010 Workshop on New Challenges
  for NLP Frameworks}.

\bibitem[{Riedl and Biemann(2012)}]{riedl2012topictiling}
Martin Riedl and Chris Biemann. 2012.
\newblock Topictiling: a text segmentation algorithm based on lda.
\newblock In \emph{Proceedings of ACL 2012 Student Research Workshop}, pages
  37--42. Association for Computational Linguistics.

\bibitem[{Rudinger et~al.(2015)Rudinger, Rastogi, Ferraro, and
  Van~Durme}]{Rudinger1025}
Rachel Rudinger, Pushpendre Rastogi, Francis Ferraro, and Benjamin Van~Durme.
  2015.
\newblock Script induction as language modeling.
\newblock In \emph{Proceedings of the 2015 Conference on Empirical Methods in
  Natural Language Processing}, pages 1681--1686.

\bibitem[{Ryu et~al.(2010)Ryu, Jung, Kim, and Myaeng}]{ryu2010automatic}
Jihee Ryu, Yuchul Jung, Kyung-min Kim, and Sung~Hyon Myaeng. 2010.
\newblock Automatic extraction of human activity knowledge from
  method-describing web articles.
\newblock In \emph{Proceedings of the 1st Workshop on Automated Knowledge Base
  Construction}, page~16. Citeseer.

\bibitem[{Salton and McGill(1986)}]{salton1986introduction}
Gerard Salton and Michael~J McGill. 1986.
\newblock Introduction to modern information retrieval.

\bibitem[{Schank and Abelson(1977)}]{schank1977}
Roger~C. Schank and Robert~P. Abelson. 1977.
\newblock Scripts, plans, goals and understanding, an inquiry into human
  knowledge structures.
\newblock \emph{Hillsdale: Lawrence Erlbaum Associates,}, 3(2):211 -- 217.

\bibitem[{Singh et~al.(2002)Singh, Lin, Mueller, Lim, Perkins, and
  Zhu}]{omics2002}
Push Singh, Thomas Lin, Erik~T Mueller, Grace Lim, Travell Perkins, and Wan~Li
  Zhu. 2002.
\newblock Open mind common sense: Knowledge acquisition from the general
  public.
\newblock In Robert Meersman and Zahir Tari, editors, \emph{On the Move to
  Meaningful Internet Systems 2002: CoopIS, DOA, and ODBASE}, pages 1223--1237.
  Springer, Berlin / Heidelberg, Germany.

\bibitem[{Tandon et~al.(2014)Tandon, de~Melo, Suchanek, and
  Weikum}]{webchild14}
Niket Tandon, Gerard de~Melo, Fabian~M. Suchanek, and Gerhard Weikum. 2014.
\newblock Webchild: harvesting and organizing commonsense knowledge from the
  web.
\newblock In \emph{Seventh {ACM} International Conference on Web Search and
  Data Mining, {WSDM} 2014, New York, NY, USA, February 24-28, 2014}, pages
  523--532.

\bibitem[{Tandon et~al.(2017)Tandon, de~Melo, and Weikum}]{tandon2017webchild}
Niket Tandon, Gerard de~Melo, and Gerhard Weikum. 2017.
\newblock Webchild 2.0: Fine-grained commonsense knowledge distillation.
\newblock \emph{Proceedings of ACL 2017, System Demonstrations}, pages
  115--120.

\bibitem[{Wanzare et~al.(2016)Wanzare, Zarcone, Thater, and
  Pinkal}]{wanzare2016descript}
Lilian D.~A. Wanzare, Alessandra Zarcone, Stefan Thater, and Manfred Pinkal.
  2016.
\newblock Descript: A crowdsourced corpus for the acquisition of high-quality
  script knowledge.
\newblock In \emph{The International Conference on Language Resources and
  Evaluation.}

\end{thebibliography}
